\UseRawInputEncoding
\documentclass{article}

\usepackage[preprint, nonatbib]{nips_2018}

\usepackage{float}

\usepackage[square,numbers]{natbib}
\usepackage[utf8]{inputenc} 
\usepackage[T1]{fontenc}    
\usepackage{hyperref}       
\usepackage{url}            
\usepackage{booktabs}       
\usepackage{amsfonts}       
\usepackage{nicefrac}       
\usepackage{microtype}      
\usepackage{graphicx}
\usepackage{tabularx}
\usepackage{hyperref}
\usepackage{amsmath}
\usepackage{doi}
\title{1st Place Solution to ICDAR 2021 RRC-ICTEXT: End-to-end Text Spotting and Aesthetic Assessment on Integrated Circuit}

\author{
  Qiyao Wang $ ^{1\ *} $  \\
  \texttt{wangqiyao@hikvision.com} \\
\And
 Pengfei Li $ ^{1 } $ \thanks{These authors  contributed  equally  to  this  work  and  should  be  considered  co-first  authors} \\
  \texttt{lipengfei27@hikvision.com}  \\
 \And
   Li Zhu $ ^{2} $\\
  \texttt{zhuli1@hikvision.com} \\
\And
 Yi Niu $ ^{3} $\\
  \texttt{niuyi@hikvision.com} \\
}

\begin{document}
\maketitle

\begin{abstract}
This paper presents our proposed methods to ICDAR 2021 Robust Reading Challenge - Integrated Circuit Text Spotting and Aesthetic Assessment (ICDAR RRC-ICTEXT 2021). For the text spotting task, we detect the characters on integrated circuit and classify them based on yolov5 detection model. We balance the lowercase and non-lowercase by using SynthText, generated data and data sampler. We adopt semi-supervised algorithm and distillation to furtherly improve the model's accuracy. For the aesthetic assessment task, we add a classification branch of 3 classes to differentiate the aesthetic classes of each character. Finally, we make model deployment to accelerate inference speed and reduce memory consumption based on NVIDIA Tensorrt. Our methods achieve 59.1 mAP on task 3.1 with 31 FPS and 306M memory (rank 1), 78.7\% F2 score on task 3.2 with 30 FPS and 306M memory (rank 1).
\end{abstract}

\section{Introduction}
Integrated Circuit Text Spotting and Aesthetic Assessment(ICTEXT) \cite{ictext} dataset is a large scale industry standard Integrated Circuit OCR dataset. The annotations consist of bounding box with a rotation degree of every character, its corresponding class(0-9, a-z, A-Z) and aesthetic classes. The objective of task 1 is to detect the location of every character and recognize it, and the objective of task 2 is to detect and recognize every character and also their respective aesthetic classes. The task 3.1 evaluates task 1 score, speed and loaded memory size, and task 3.2 evaluates task 2 score, speed and loaded memory size. The final rank is based on task 3.1 and task 3.2 separately and we get rank first both.

We treat the text spotting task as an object detection task for characters on integrated circuit. We follow the yolov5 \cite{yolov5}framework to build our task 1 pipeline, and it does very well compared to other detection framework. As the task takes mAP as the evaluation metric, the error of every class contributes the same. We notice that the APs of lowercases are much worse than non-lowercase, which is because there are much less lowercases than non-lowercase letters in the dataset. Therefore, we balance the lowercase and non-lowercase by using SynthText, generated data and data sampler. Then, we use semi-supervised algorithm to take advantage of the 3000 no-annotation samples.

As for task 2, we build the aesthetic assessment as a classification task by add a classification branch of 3 classes to yolov5. The three aesthetic classes are blurry text, low contrast text and also broken text. We use three binary classifiers to determine the three aesthetic classes.

As the task 3.1 and task 3.2 take speed and memory for consideration, we put some efforts in it. Considering speed and memory, we adopt yolov5s model, which is very small and fast. However, the performance of yolov5s is worse than yolov5x. Hence, we adopt distillation to enhance the accuracy of yolov5s and take yolov5x as teacher model. To accelerate testing speed and reduce memory consumption during inference step, we make model deployment based on NVIDIA Tensorrt \cite{tensorrt}.

The rest of the paper is organized as follows. We introduce the method of text spotting task(task 1) in section \ref{sec:task1}, aesthetic assessment(task 2) in section \ref{sec:task2}, model deployment(task 3.1\&3.2) in section \ref{sec:task3}, experimental results in section \ref{sec:experiments}. And finally section \ref{sec:conclusion} concludes the paper.

\section{Methods}
\label{sec:methods}
In this section, we describe our approach on 3 tasks included text spotting, aesthetic assessment and inference speed, model size and score assessment.

\subsection{Task 1: Text Spotting}
\label{sec:task1}

In the text spotting task, we detect the characters on integrated circuit and classify them at the same stage based on yolov5 detection model. Our approach mainly use those strategies to improve mAP.

\textbf{Pre-training data}. We random sample 80k data from Synthtext\cite{gupta2016synthetic} as pre-training data.

\textbf{RFS}. Repeat Factor Sampling\cite{gupta2019lvis} is adopted.

\textbf{Data Augmentation}. Mosaic~\cite{bochkovskiy2020yolov4}, hist equalization, random rotate 0 or 90 or 180 or 270 degrees are mainly used.

\begin{figure}[ht]
	\centering
	\includegraphics[scale=0.2]{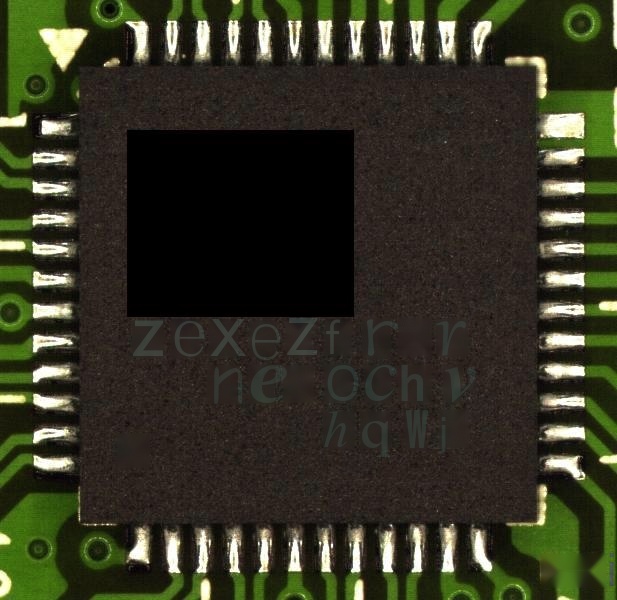}
	\includegraphics[scale=0.2]{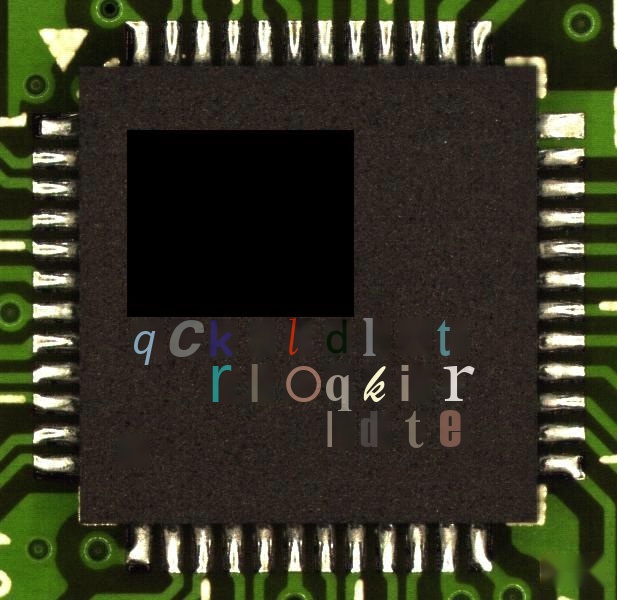}
	\caption{examples by generating lower characters using windows fonts.}
	\label{examples 1}
\end{figure}

\begin{figure}[ht]
	\centering
	\includegraphics[scale=0.3]{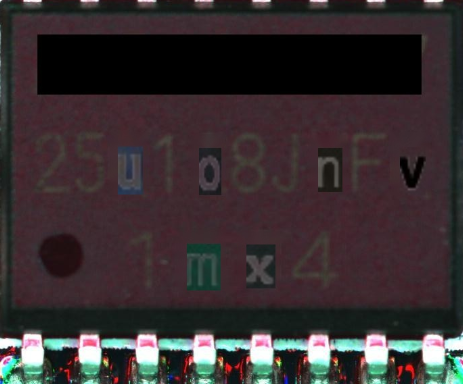}
	\caption{examples by style transforing in style text way.}
	\label{examples 2}
\end{figure}

\textbf{Synthetic data generation}. To balance the lowercase classes and non-lowercase classes, we get 2.4k samples by generating lowercase with white color and windows fonts to replace the text area of training data, 1k samples by putting the colorful lowercase generated by windows fonts on the text area of training data and 1k samples by style transference in style text~\cite{wu2019editing} way. Examples are shown at figure \ref{examples 1} and \ref{examples 2}.


\textbf{Semi supervisid learning}. We use WBF~\cite{solovyev2019weighted} to get pseudo labels on validation data.

\textbf{Iou aware}. We use iou aware\cite{wu2020iou} on training stage. We use the iou between model's prediction bounding box and ground truth bounding box as the ground truth of bounding box's confidence instead of ciou.

\subsection{Task 2: Aesthetic Assessment}
\label{sec:task2}

We build the aesthetic assessment as a classification task by add a classification branch of 3 classes to yolov5. We use a 0-1 vector of length 3 to describe the aesthetic classes of each character, the same as the annotations. In task 1, the length of vector that each anchor outputs is (n+5), where n is 62(62 classes, which is 0-9, a-z, A-Z), and 5 is [prob, x, y, w, h]. In task 2, the length of vector becomes to be (n+8), adding the aesthetic vector of length 3.

In the training step, we use Binary Cross Entropy Loss (BCE loss) to compute the aesthetic classification loss. Moreover, in the testing step, we take a threshold to determine to output 0 or 1 based on the aesthetic vector output. In our approach, we take 0.2 as threshold for three classes.

\subsection{Task 3.1 \& Task 3.2: Inference Speed, Model Size and Score Assessment}
\label{sec:task3}

The evaluation metric of task 3.1 and task 3.2 is \ref{task3.1score}:
\begin{equation}
	\label{task3.1score}
	3S=(0.2\times normalised\_speed)+(0.2\times (1-normalised\_size))+(0.6\times normalised\_score);
\end{equation}

where normalised\_speed and normalised\_size are \ref{normalisedspeed}:
\begin{subequations}
	
\label{normalisedspeed}
\begin{align}
normalised\_speed&=min(\frac{actual\_FPS}{acceptable\_FPS}, 1); \\
normalised\_size&=min(\frac{allocated\_memory\_size}{acceptable\_memory\_size}, 1);
\end{align}
\end{subequations}
where acceptable\_FPS is 30FPS and acceptable\_memory\_size is 4000MB;

The normalised\_score for task 3.1 is evaluated as task 1 and for task 3.2 is evaluated as task 2. Therefore, we can get full score for speed if our FPS > 30 and we need to make allocated memory as less as possible.

To improve the final score, we adopt these strategies as below.

\textbf{Smaller model}. Based on task 1's strategies, we retrain the smaller model to get lower GPU memory and faster inference speed.

\textbf{Knowledge Distillation}. To improve the smaller model's performance, we apply knowledge distillation on it. We make several modification on standard knowledge distillation\cite{hinton2015distilling}: (1)We split the distillation feature to classifier feature and itself, and use MSE loss on distillation feature, KL loss on classifier feature. (2)We apply a mask on the text area of rare classes and iou between teacher model's box prediction and ground truth that is bigger than 0.5. The motivation is that we force student model only to learn the best feature from teacher and we find that the rare classes's performance gap between teacher and student is bigger. The formula details are shown in \ref{distillation equation}. Where $ f_t $, $ f_s $ represent feature of teacher and student, $ f_{t-cls} $, $ f_{s-cls} $ represent classifier feature.

\begin{subequations}
	\label{distillation equation}
	\begin{align}
		loss_{distillation} &= MSE(f_{t}*mask, f_{s}*mask) + KL(f_{t-cls}, f_{s-cls}) \\
		mask[i,j] &= \begin{cases}
			1, & if \quad iou(t_{box}, gt_{box})>0.5 \quad and \quad class_{gt[i,j]} \in rare-classes \\
			0, & otherwise
		\end{cases}
	\end{align}
\end{subequations}

\textbf{Deployment}. We use NVIDIA Tensorrt to deploy our pytorch model. The model deployment can significantly accelerate inference speed and reduce memory consumption. To balance score between speed and size, we set input image size to 672*672, batchsize to 1, box confidence threshold to 0.01 and NMS threshold to 0.1.

\section{Experimental results}
\label{sec:experiments}
\subsection{Dataset}
We perform experiments on ICTEXT dataset, which contains 7k training data and 3k validation data. Note that we can't get the annotations of validation data, so we submit our methods on task 1 to get the experiments' results.

\subsection{Implementation Details}

We use yolov5 as our method. All the hyper-parameters are kept unchanged except some data augmentation. The batch size is 16 at single GPU and the training image size is set to 736. We consider the lowercase classes as rare classes. Firstly, we train yolov5x from scratch on pre-training data, then we finetune our model on training data and synthetic data. After that, we use yolov5x to predict the validation data rotated by 0, 90 180, 270 degrees, and use WBF to make pseudo labels. Finally,
we train yolov5x as the teacher model by training data, synthetic data and validation data with pseudo labels, and train yolov5s as the student model. Besides,
we use knowledge distillation to improve yolov5s' performance. we choose the models' output as distillation feature and set the distillation loss weight as 0.05.

\subsection{Ablation Studies}

We choose yolov5x as our baseline model, Some useful enhancement techniques are shown in Table \ref{Ablation studies on validation data}. with those methods, we improve the mAP from 0.54 to 0.60. Based on those techniques, we train yolov5s and use knowledge distillation to further improve the mAP. the results are shown in Table \ref{Yolov5s'ablation studies}.
\begin{table}[!htb]
	\centering
	
	\label{Ablation studies on validation data}
	\resizebox{0.7\textwidth}{!}{%
		\begin{tabular}{ccccc|ccc}
			\toprule
			PT          & RFS & DA & SG & SSL &   AP & AP@0.5 & AP@0.75 \\
			\midrule
			            &            &   &   &   &  54.2  &  71.7 & 65.5\\
			\checkmark  &           &   &  &  &  57.0&  73.5  &  69.0 \\
			\checkmark & \checkmark &   &  &  &  57.4  & 75.3   &  69.3  \\
			\checkmark & \checkmark & \checkmark &  &  &   57.8  & 74.7   &  70.4 \\
			\checkmark & \checkmark & \checkmark & \checkmark & & 59.6 & 77.4  & 72.0   \\
			\checkmark & \checkmark & \checkmark & \checkmark & \checkmark & \textbf{60.4}   & \textbf{78.0}  &  \textbf{72.8}\\
			\bottomrule
	\end{tabular}}
	\caption{Ablation studies on validation data. PT:pre-training;RFS:Repeat Factor Sampling;DA: Added data augmentation compared with the default setting:hist equalization,rotate etc;SG:Synthetic data generation;SSL:Semi supervisid learning}
\end{table}

\begin{table}[!htb]
	\centering
	
	\label{Yolov5s'ablation studies}
	\resizebox{0.5\textwidth}{!}{%
		\begin{tabular}{cccc}
			\toprule
			 & AP & AP@0.5 & AP@0.75 \\
			\midrule
			Baseline   &  57.4   &   74.0    &  69.3     \\
			SKD  & 58.5  &75.2  &   70.2     \\
			\textbf{MKD}  & \textbf{59.4}  &\textbf{76.2}  &   \textbf{70.9}   \\
			\bottomrule
	\end{tabular}}
	\caption{Yolov5s'ablation studies on validation data. Baseline:using tabel 1's techniques, SKD:standard knowledge distillation, MKD:knowledge distillation with mask}
\end{table}

\newpage

\subsection{Final Results}
We submit our docker image on task 3 evaluation server, and the final results are shown on Table 2 and Table 3. The yolov5x model performs better than yolov5s but it takes a lot longer inference time and allocates more GPU memory, so the 3S score of yolov5s is higher. Finally, we deploy our yolov5s model by NVIDIA Tensorrt and make a big progress both on task 3.1 and task 3.2.

\begin{table}[!htb]
  \label{table3.1}
  \centering

  \begin{tabular}{ccccccc}
    \toprule
    Name    & AP    & AP IOU@0.5    & AP IOU@0.5    & FPS    & GPU Memory(MB)    & \textbf{3S} \\
    \midrule
    yolov5x(pytorch)    & 0.62    & 0.79    & 0.76    & 3.69    & 1323.75    & 0.53 \\
    yolov5s(pytorch)    & 0.59    & 0.76    & 0.72    & 25.45    & 745.75    & 0.69 \\
    \textbf{yolov5s(tensorrt)}    & 0.59    & 0.76    & 0.73    & 31.04    & 305.88    & \textbf{0.74} \\
    \bottomrule
  \end{tabular}
  \caption{Final results on task 3.1}
\end{table}

\begin{table}[!htb]
  \label{table3.2}
  \centering

  \begin{tabular}{ccccccc}
    \toprule
    Name    & Multi-Label    & Multi-Label    & Multi-Label    & FPS    & GPU    & \textbf{3S} \\
            &Precision       & Recall         & F-2 Score &  & Memory(MB)  &  \\
    \midrule
    yolov5s(pytorch)    & 0.77    & 0.77    & 0.78    & 25.44    & 609.75    & 0.81 \\
    \textbf{yolov5s(tensorrt)}    & 0.77    & 0.77    & 0.79    & 29.68    & 305.88    & \textbf{0.85} \\
    \bottomrule
  \end{tabular}
  \caption{Final results on task 3.2}
\end{table}

\section{Conclusion}
\label{sec:conclusion}
In this paper, we introduced our 1st place methods to ICDAR 2021 RRC-ICTEXT on the whole tasks. We built text spotting pipeline based on yolov5 model to detect single characters and classify them at the same time. We added a classification branch to the detection model to do aesthetic assessment while text spotting. The whole pipeline is quite simplified without any redundant process and got rank 1st in task 1 and task 2. Finally, we deployed the model to make it faster and load less memory and got 0.74(3S score) for task 3.1 and 0.85(3S score) for task 3.2, which are ranked 1st on both tasks.

\bibliography{references.bib}

\begin{thebibliography}{10}
\providecommand{\natexlab}[1]{#1}
\providecommand{\url}[1]{\texttt{#1}}
\expandafter\ifx\csname urlstyle\endcsname\relax
  \providecommand{\doi}[1]{doi: #1}\else
  \providecommand{\doi}{doi: \begingroup \urlstyle{rm}\Url}\fi

\bibitem[ict()]{ictext}
Icdar rrc-ictext.
\newblock URL \url{https://ictext.v-one.my}.

\bibitem[ten()]{tensorrt}
tensorrt.
\newblock URL \url{developer.nvidia.com}.

\bibitem[yol()]{yolov5}
yolov5.
\newblock URL \url{https://github.com/ultralytics/yolov5}.

\bibitem[Bochkovskiy et~al.(2020)Bochkovskiy, Wang, and
  Liao]{bochkovskiy2020yolov4}
A.~Bochkovskiy, C.-Y. Wang, and H.-Y.~M. Liao.
\newblock Yolov4: Optimal speed and accuracy of object detection.
\newblock \emph{arXiv preprint arXiv:2004.10934}, 2020.

\bibitem[Gupta et~al.(2016)Gupta, Vedaldi, and Zisserman]{gupta2016synthetic}
A.~Gupta, A.~Vedaldi, and A.~Zisserman.
\newblock Synthetic data for text localisation in natural images.
\newblock In \emph{Proceedings of the IEEE conference on computer vision and
  pattern recognition}, pages 2315--2324, 2016.

\bibitem[Gupta et~al.(2019)Gupta, Dollar, and Girshick]{gupta2019lvis}
A.~Gupta, P.~Dollar, and R.~Girshick.
\newblock Lvis: A dataset for large vocabulary instance segmentation.
\newblock In \emph{Proceedings of the IEEE/CVF Conference on Computer Vision
  and Pattern Recognition}, pages 5356--5364, 2019.

\bibitem[Hinton et~al.(2015)Hinton, Vinyals, and Dean]{hinton2015distilling}
G.~Hinton, O.~Vinyals, and J.~Dean.
\newblock Distilling the knowledge in a neural network.
\newblock \emph{arXiv preprint arXiv:1503.02531}, 2015.

\bibitem[Solovyev et~al.(2019)Solovyev, Wang, and
  Gabruseva]{solovyev2019weighted}
R.~Solovyev, W.~Wang, and T.~Gabruseva.
\newblock Weighted boxes fusion: ensembling boxes for object detection models.
\newblock \emph{arXiv preprint arXiv:1910.13302}, 2019.

\bibitem[Wu et~al.(2019)Wu, Zhang, Liu, Han, Liu, Ding, and Bai]{wu2019editing}
L.~Wu, C.~Zhang, J.~Liu, J.~Han, J.~Liu, E.~Ding, and X.~Bai.
\newblock Editing text in the wild.
\newblock In \emph{Proceedings of the 27th ACM international conference on
  multimedia}, pages 1500--1508, 2019.

\bibitem[Wu et~al.(2020)Wu, Li, and Wang]{wu2020iou}
S.~Wu, X.~Li, and X.~Wang.
\newblock Iou-aware single-stage object detector for accurate localization.
\newblock \emph{Image and Vision Computing}, 97:\penalty0 103911, 2020.

\end{thebibliography}

\end{document}